\documentclass[journal]{IEEEtai}

\usepackage[colorlinks,urlcolor=blue,linkcolor=blue,citecolor=blue]{hyperref}

\usepackage{color,array}
\usepackage{xcolor}

\usepackage{graphicx}
\usepackage[numbers]{natbib}

\usepackage{multirow}

\begin{document}

\title{A Review of Text Style Transfer using Deep Learning}

\author{Martina Toshevska and Sonja Gievska \\
Faculty of Computer Science and Engineering, Ss. Cyril and Methodius University - Skopje, \\
North Macedonia\\
Email: \{martina.toshevska, sonja.gievska\}@finki.ukim.mk
}


\maketitle

\begin{abstract}
Style is an integral component of a sentence indicated by the choice of words a person makes. Different people have different ways of expressing themselves, however, they adjust their speaking and writing style to a social context, an audience, an interlocutor or the formality of an occasion. Text style transfer is defined as a task of adapting and/or changing the stylistic manner in which a sentence is written, while preserving the meaning of the original sentence.

A systematic review of text style transfer methodologies using deep learning is presented in this paper. We point out the technological advances in deep neural networks that have been the driving force behind current successes in the fields of natural language understanding and generation. The review is structured around two key stages in the text style transfer process, namely, representation learning and sentence generation in a new style. The discussion highlights the commonalities and differences between proposed solutions as well as challenges and opportunities that are expected to direct and foster further research in the field.
\end{abstract}

\begin{IEEEImpStatement}
Motivated by recent advancements in the field, we have carried out a systematic review of state-of-the-art research to highlight the trends, commonalities and differences across style transfer methodologies using deep learning. The discussion is organized around key stages of the process, namely, representation learning of style and content of a given sentence, and generation of the sentence in a new style. A comprehensive view of methodologies, available datasets and evaluation metrics is compiled to foster further research in the field.
\end{IEEEImpStatement}


\section{Introduction}

Naturally occurring linguistic variations in spoken and written language have been contributed to culture, personal attributes and social context~\cite{eckert2001style, coupland2007style}. The underlying factors contributing to linguistic variations in spoken language have been extensively studied in the field of variationist sociolinguistics. The adjustments of one’s individual style to match or shift away~\cite{kiesling1998language} from the style of the interlocutor, the audience or social context are prominent in the work of the American linguist, William Labov~\cite{labov1972sociolinguistic, labov1981field, labov1972some}. Different people have different ways of expressing themselves~\cite{labov1972sociolinguistic} and personal attributes, such as gender, age, education, personality, emotional state~\cite{pennebaker2003psychological} are reflected in their writing style. However, style changes over time~\cite{eckert2001style} and we adjust to a social context, an audience we address, a person we communicate with~\cite{bell1984language}, and/or the formality of an occasion~\cite{kiesling1998language}. While direct mapping of sociolinguistics categories is not always possible, stylistic properties have been classified along several dimensions in the research on natural language understanding and generation.

Adjusting the style of a sentence by rewriting the original sentence in a new style, while preserving its semantic content, is referred to as text style transfer. The diversity of linguistic styles is matched by the diversity in research interests in the field. Some researchers viewed style transfer as an ability to adjust the emotional content in a written text; others equated the concept with formality or politeness. Changing the sentiment polarity of a sentence might change the meaning of a text or transform the message it conveys, although the ability to change the emotional content in a written text should be viewed more along the lines of adjusting the tone of a message that is more appropriate, emphatic and less severe or offensive to the audience or the conversational partner. Other researchers have directed their efforts towards much more sound conceptualization of a style as a genre, or linguistic style of a person, or a particular social group. 

Language style should be a special consideration in current and future intelligent interaction systems~\cite{li2016persona} that understand, process, or generate speech or text. Automatically adjusting the text style could help users improve their communication skills (e.g., being more polite, learning to write formal messages), and could become even more important, when employed in future prosocial interaction mediators on discussion platforms and comment-based communities (e.g., toning down negative sentiment, neutralizing offensiveness).

In a decade or so, the work on the topic expanded from a few articles to an active research area. Most of the methods for text style transfer are based on deep neural networks. The success of deep learning in other areas has provided fruitful directions to be followed. Inspired by the success of the encoder-decoder models in other fields, including machine translation (MT)~\cite{cho2014properties, wu2016google}, text summarization~\cite{chopra2016abstractive} and dialogue generation~\cite{serban2017hierarchical}, a number of style transfer models are built upon this end-to-end model of learning~\cite{li2018delete, sudhakar2019transforming, xu2019formality, dai2019style, cheng2020contextual, fu2018style, dos2018fighting, prabhumoye2018style, john2019disentangled, zhang2018learning, prabhumoye2018style2, tian2018structured, liu2020revision, wang2019controllable, zhou2020exploring, lee2020stable, zhang2018shaped, hu2017toward}. New advances directed toward adversarial learning have also inspired more recent works on text style transfer~\cite{shen2017style, zhao2018language, logeswaran2018content, chen2018adversarial}.

Pivotal in this review are the studies that address the automatic adjustment of the style of a written text. At the onset of our paper, we introduce the reader to various text styles that have been in the focus of the selected research papers. We have compiled a list of publicly available datasets that we discuss in terms of their suitability for a particular style transfer task(s). The evaluation of how successful a particular model is on the task of style transfer has two objectives: to measure how well the semantic content in the generated sentence was preserved and to assess the quality of rewriting the sentence in a new (target) style. Evaluation of the performance of style transfer models is of special importance for future research in the area. 

The discussion of the specifics of the proposed approaches to style transfer is organized to allow readers to follow the advances in deep learning and their impact on style transfer tasks. The discussion follows the two key stages in the text style transfer process: 1) representation learning of the style and content of a given sentence and 2) generation of the sentence that has the same meaning as the input sentence, but is expressed in a different style. Auxiliary elements, such as style embeddings~\cite{li2018delete, sudhakar2019transforming, dai2019style, cheng2020contextual, fu2018style, dos2018fighting, zhang2018learning, lee2020stable, hu2017toward, shen2017style, logeswaran2018content, chen2018adversarial}, style classifiers~\cite{xu2019formality, dai2019style} and/or adversarial discriminators~\cite{shen2017style, zhao2018language} are also discussed. We discuss the critical stage of the process, the output sentence generation by categorizing the approaches among three groups. Namely, models that use a simple approach to generation by reconstructing the input sentence, models that incorporate additional style classifier in their encoder-decoder architectures, and models that adopt adversarial learning. 

The paper is organized as follows. After the introductory section, Section~\ref{sec:textual_style} provides a description of various text styles that have been in the focus of the selected research papers. Section~\ref{sec:style_transfer_for_text} gives a formalization of text style transfer and discusses the publicly available datasets suitable for the task at hand. The discussion of a set of measures, which have been proposed as meaningful criteria for evaluating style transfer models, is also presented. Beginning with a brief introduction of several deep neural networks in Section~\ref{sec:deep_neural_networks_for_text_generation}, the discussion of state-of-the-art style transfer methodologies using deep learning is presented in Section~\ref{sec:methods_for_style_transfer}. Section~\ref{sec:challenges} reflects on the challenges style transfer faces and casts light on potential research directions that are expected to further advance the field. Section~\ref{sec:conclusion} concludes the paper.


\section{Text Style}
\label{sec:textual_style}

The nuance and subtilty of language variations are functions of individual, social as well as situational differences. Emerging research on automatic style transfer of written text converges toward a common view that style is an integral part of a sentence, indicated by the choices of words a person makes~\cite{argamon2010rest}. We provide an introduction of various text styles that have been in the focus of the research on the automatic style transfer. In particular, a short description of the following linguistic styles is given: individual style, genre, as well as the formality, politeness, offensiveness, and sentiment that is conveyed by a given text. 

\subsection{Personal Style}

The words people use reveal a lot about themselves, such as their personality, gender, or age~\cite{pennebaker2003psychological}. There are several studies examining language variations across different gender and age groups found in formal texts~\cite{argamon2003gender}, social media~\cite{peersmaneffects, bamman2014gender}, and blog posts~\cite{koppel2002automatically, schler2006effects}. The findings suggest that differences in language usage between various demographic groups do exist and the identification of the author’s gender and/or age could be done with an accuracy of 80\% on the basis of usage of specific words~\cite{koppel2001automatically}. For instance, female users tend to use more emoticons~\cite{rao2010classifying} and choose words with positive emotional connotation~\cite{preotiuc2016discovering}. On the other hand, the study of online language highlights the differences between various age groups - younger people use chat-specific e-language and refer to themselves more frequently, while older people use more complex sentences and include more links and hashtags~\cite{nguyen2013old}.

These findings could be fruitfully applied in human-computer interaction~\cite{li2016persona} as important interaction features that develop user trust and satisfaction. A key challenge in designing believable virtual assistants is endowing them with dialog capabilities that are not only responsive to the user's need, but have a style of their own that matches the user's language style. 

Shakespearean writing style\footnote{https://en.wikipedia.org/wiki/Shakespeare's\_writing\_style, last visited: 09.03.2021} has been recognized as a specific writing style. An interesting research task of rewriting sentences in Shakespearean style have been reported in~\cite{jhamtani2017shakespearizing} that could be potentially used for edutainment purposes. Research on generating image caption used a rather unorthodox approach to generating caption in a style that was learned from romance novels and Taylor Swift’s song lyrics\footnote{https://medium.com/@samim/generating-stories-about-images-d163ba41e4ed, last visited: 01.04.2021}~\cite{kiros2015skip, zhu2015aligning}.

\subsection{Formality}

Language style is very often associated with register i.e., formality of a given text. There is no unified definition of what formal language is and yet, the distinction between the language used in formal and informal settings is well recognized. For example, the language in academic papers is considered more formal than the language used in social media. Longer texts as well as texts containing passive voice tend to be perceived as more formal~\cite{sheikha2010automatic}. The formal style of writing is usually characterized by detachment, precision, objectivity, rigidity, and higher cognitive load~\cite{heylighen1999formality}. On the other side, texts that contain short words, contractions, and abbreviations are considered informal~\cite{sheikha2010automatic}. The informal style is more subjective, less accurate, less informative, and with a much lighter form~\cite{heylighen1999formality}. Indicators of formality considered in the research on automatic formality detection include the use of slang and grammatically incorrect words~\cite{peterson2011email}, social distance, and shared knowledge between the writer and the audience~\cite{lahiri2011informality}. Automatically improving the level of formality of a written text is a useful feature incorporated in writing assistants~\cite{rao2018dear}.

\subsection{Politeness}

The politeness of the language we use is affected by the social distance between the writer and the audience~\cite{brown1987politeness, chilton1990politeness, danescu2013computational}. The level of politeness is important for “maintaining a positive face” in social interaction with others~\cite{coppock2005politeness} and it plays a significant role in the overall experience of communication~\cite{andersson1999tit}. Polite and impolite are located on the opposite sides of the spectrum, although different levels of politeness might be used. A study presented in~\cite{chhaya2018frustrated}, shows that high frustration is correlated with a writing style that is less polite and less formal. Systems for automatic adjustment of politeness could safeguard online writing, especially in a situation when someone (unintentionally) writes an impolite text that will be received and read by others.

\subsection{Offensiveness}

The damaging consequences of malicious online behavior in the form of hate speech, trolling, and use of offensive language remain a recurrent problem for almost any social media platform. Devising systems and establishing interaction mediators that will automatically identify, remove, and/or label posts with offensive language and hate speech is demanded by public, governments, and institutions. 

Detecting offensive language is a widespread research area that focuses on determining whether a sentence is offensive or not~\cite{pavlopoulos2019convai}, or determining the audience that is targeted by a message (group or individual)~\cite{zampieri2019predicting}. Studies show that usage of specific words might correlate with offensive language. For example, words, such as "killed", "fool", "ignorant" are often correlated with offensive language~\cite{risch2020offensive}. The potential benefit of a style transfer system to neutralize offensive remarks before they are posted is welcomed by many social media and comment-based news communities.

\subsection{Genre}

Genre of a document is determined on the basis of some external criteria~\cite{lee2001genres}, such as purpose and target audience~\cite{biber1991variation}. News, advertisements, and technical reports are some of the genres text documents are categorized into. Identifying the genre of a document could potentially improve Information Retrieval systems by search results that match or are relevant to a particular user’s search. For example, when one intends to buy something, advertisements might be more relevant than scientific reports~\cite{dewdney2001form}. 

A document written in a style that matches the language style used by a particular group of people is expected to be more understandable by the target audience. For instance, medical reports are often difficult to understand by non-experts in the field. Automatically transforming a medical report into a document in layman terms, might improve its readability by a wider audience.

\subsection{Sentiment}

Emotions play a crucial role in human behavior~\cite{baumeister2007emotion} and one’s emotional state is often reflected in one's spoken or written language. While emotional connotation carried by a sentence may not be a typical stylistic variation of language, rewriting a sentence with toned down negative emotions might be desired in many applications. Several categorical and dimensional models for emotions have been proposed~\cite{russell1980circumplex, plutchik1984emotions, ekman1992argument, susanto2020hourglass}. Detecting sentiment polarity of a text i.e. whether the overall sentiment of a particular text is positive or negative~\cite{pang2002thumbs, kim2004determining, akhtar2020intense, delbrouck2020transformer, javdan2020applying, shenoy2020multilogue} have been used in predictive analytics. Being able to detect emotions expressed in online posts have been used to “sense the mood of a community”~\cite{de2013predicting, hasan2014emotex, ma2018targeted, kale2021fragmented}, opinion of the public about specific events~\cite{vo2013twitter}, the emotions embedded in news headlines~\cite{strapparava2008learning}, or political sentiment~\cite{khatua2020predicting}. Sentiment polarity has been used for predicting the impact of users’ reviews on book sales~\cite{chevalier2006effect}, sales performance prediction~\cite{liu2007arsa}, ranking products based on user reviews~\cite{mcglohon2010star}, stock market prediction according to Twitter moods~\cite{bollen2011twitter}, website popularity prediction~\cite{li2020popularity}, etc.

\section{Style Transfer for Text}
\label{sec:style_transfer_for_text}


\subsection{Style Transfer Tasks}

Text style transfer refers to the process of rewriting a sentence in a new style, which involves generating a new (output) sentence that has the same explicit meaning as the original (input sentence), while stylistically differing from the original one. Style transfer has been applied to adjust, modify or adapt the manner in which a sentence is written. The term style has been used rather broadly and encompasses properties, such as: \textit{register (formality)}, \textit{politeness}, \textit{offensiveness}, \textit{genre according to purpose}, \textit{genre according to the target audience}, \textit{sentiment} or the \textit{individual style of the author or the social group they belong to}. Table~\ref{tab:tasks} presents illustrative examples for each of the style transfer tasks that have been given attention in research literature.

The objective of each style transfer task is to adjust the style of a sentence with respect to particular style properties. For example, adjusting the emotions conveyed in a sentence is referred to as \textit{sentiment style transfer}. Adjusting the politeness or the formality of a sentence is associated with \textit{politeness} and \textit{formality transfer}, respectively. Removing the offensiveness and substituting it with a neutral style has been the objective in the task of \textit{transferring offensive to non-offensive text}. Rewriting a text that stylistically adheres to the personal writing style of an author (e.g., Shakespeare writing style, Taylor Swift’s lyrics) or a social group (e.g., masculine vs feminine language style, democrat vs republican language) is referred to as \textit{personal style transfer}. \textit{Genre style transfer} could be related to a purpose (i.e. advertisement or news articles are written in a different style) or the intended audience (e.g., content written in expert language vs layman language).


\begin{table*}
\centering
\begin{tabular}{lll}
\hline
\textbf{Task} & \textbf{Input sentence (style 1)} & \textbf{Output sentence (style 2)} \\
\hline
Sentiment style transfer & \shortstack[l]{\textit{Great food, but horrible staff and very} \\ \textit{very rude workers!} (negative)} & \shortstack[l]{\textit{Great food, awesome staff, very personable} \\ \textit{and very efficient atmosphere!} (positive)} \\ \hline
Politeness transfer & \textit{Send me the data.} (non-polite) & \textit{Could you please send me the data?} (polite) \\ \hline
Formality transfer & \textit{Gotta see both sides of the story.} (informal) & \textit{You have to consider both sides of the story.} (formal) \\ \hline
Transferring offensive to non-offensive text & \shortstack[l]{\textit{I hope they pay out the ***,} \\ \textit{fraudulent or no.} (offensive)} & \shortstack[l]{\textit{I hope they pay out the state,} \\ \textit{fraudulent or no.} (non-offensive)} \\ \hline
Personal style transfer (Shakespearean) & \shortstack[l]{\textit{My lord, the queen would speak with you,} \\ \textit{and presently.} (shakespearean english)} & \shortstack[l]{\textit{My lord, the queen wants to speak with} \\ \textit{you right away.} (contemporary english)} \\ \hline
Genre based on audience (expert/layman) & \shortstack[l]{\textit{Many cause dyspnea, pleuritic chest pain,} \\ \textit{or both.} (expert)} & \shortstack[l]{\textit{The most common symptoms, regardless of the} \\ \textit{type of fluid in the pleural space or its cause,} \\ \textit{are shortness of breath and chest pain.} (layman)} \\
\hline
\end{tabular}
\caption{\label{tab:tasks} Illustrative examples of selected style transfer tasks.}
\end{table*}

\subsection{Datasets}

A number of datasets have been used in the research on style transfer of text. The list of publicly available datasets targeted by the research on style transfer offered in this paper is presented in Table~\ref{tab:datasets}. Each dataset has been described with the following attributes: the year when a dataset has been published, whether the dataset is composed of parallel text sample pairs or not, type of text (e.g., emails, reviews, tweets, posts, documents), the number of text data samples, labels for the style used, as well as references to studies that have previously used the dataset.

A short description of the datasets, divided into parallel and non-parallel is presented below. The number of parallel datasets suitable for style transfer is limited since creating a large number of pairs of text (e.g., sentences, paragraphs, documents) containing sentences that express the same meaning in a different manner requires a lot of human work. In non-parallel datasets there are no paired data to learn from. The number of these datasets is larger because most of them are subsets or adapted versions of datasets that have been previously created for other tasks, such as sentiment analysis, author profiling, genre classification, etc.

\subsubsection{Parallel Datasets}

\textbf{Shakespeare}\footnote{https://github.com/cocoxu/Shakespeare, last visited: 28.08.2020} dataset~\cite{xu2012paraphrasing, xu2014data} contains 21,075 sentence pairs from 16 Shakespeare's plays and their line-by-line paraphrases in contemporary English. A style transfer task on this dataset has been defined as a transformation of a sentence written in contemporary English into a sentence written in Shakespeare’s language style.

\textbf{GYAFC}\footnote{https://github.com/raosudha89/GYAFC-corpus, last visited: 28.08.2020} dataset (Grammarly's Yahoo Answers Formality Corpus)~\cite{rao2018dear} is a parallel dataset of formal and informal sentence pairs. A subset of informal sentences is selected from the Yahoo Answers L6 corpus\footnote{https://webscope.sandbox.yahoo.com/catalog.php?datatype=l, last visited: 28.08.2020}. For each sentence, a formal version is written by people recruited through Amazon Mechanical Turk (AMT). The final dataset contains 112,975 pairs of informal-formal sentences and 111,266 pairs of formal-informal sentences.

\citet{cheng2020contextual} have created a dataset containing informal and formal versions of 600,000 email messages from the \textbf{Enron corpus}~\cite{klimt2004introducing}. The AMT annotators were asked to identify informal sentences in each email and rewrite them in a formal style.

\textbf{Captions}~\cite{gan2017stylenet} is composed of 7,000 image captions that were classified as factual, romantic, or humorous. For each image, a caption in all three styles is created, making the dataset appropriate for a stylistic transformation of a sentence among the three alternate styles.

\subsubsection{Non-parallel Datasets}

\textbf{Yelp}\footnote{https://www.yelp.com/dataset, last visited: 13.03.2020} dataset is a collection of 8.6 million business reviews that are classified as positive or negative according to their 5-star rating system, making the dataset suitable for sentiment style transfer. The dataset was often used to train systems for changing the polarity of a given text.

\textbf{Gender}~\cite{reddy2016obfuscating} is a subset of Yelp reviews annotated with gender labels (male and female) that were assigned by inferring the gender of the review's author by his or her first name. This subset is suitable for rewriting text written by a female in a masculine writing style (and vice versa).

\textbf{Amazon}\footnote{http://jmcauley.ucsd.edu/data/amazon/, last visited: 13.03.2020} dataset~\cite{he2016ups} of 1 million product reviews, \textbf{SST}\footnote{https://nlp.stanford.edu/sentiment/treebank.html, last visited: 27.08.2020} (Stanford Sentiment Treebank)~\cite{socher2013recursive} consisting of 9,613 movie reviews, and \textbf{IMDB}~\cite{diao2014jointly} dataset composed of 350,000 movie reviews have been labeled with sentiment polarity making them often used for sentiment transfer.

\textbf{Paper-News Titles} dataset~\cite{fu2018style} contains 200,000 titles categorized into two groups: titles of scientific articles and headlines of news articles. The news are collected from the UC Irvine Machine Learning Repository and the papers were compiled from publishing websites, such as ACM Digital Library, Springer, Nature, Science Direct, arXiv, and others.

\textbf{Gigaword dataset}~\cite{graff2003english, napoles2012annotated} is composed of 4 million news articles from seven news media publishers. The headlines of the news articles have been labeled according to their publisher.

\textbf{Political slant}~\cite{voigt2018rtgender} is a dataset composed of 540,000 comments on Facebook posts from 412 members of the United States Senate and House of Representatives. Every comment is categorized as either democratic or republican on the basis of the political affiliation of a member.

\citet{dos2018fighting} have created two datasets, \textbf{Twitter} containing 2 million tweets and \textbf{Reddit} dataset of 7.5 million sentences. 
The datasets have been used in research on style transfer i.e. adjusting or removing the offensiveness in a sentence.

\citet{madaan2020politeness} use the Enron corpus~\cite{klimt2004introducing} to create \textbf{Politeness} dataset\footnote{https://github.com/tag-and-generate/politeness-dataset, last visited: 28.08.2020} of 270,000 emails, labeled as polite or impolite. The potential use of this dataset would be to convert a neutral sentence into a polite sentence.

\citet{cao2020expertise} have created the \textbf{Expertise} dataset\footnote{https://srhthu.github.io/expertise-style-transfer/\#disclaimer, last visited: 26.02.2021} based on the Mericks Manuals that is suitable for transfer of style between expert and layman medical language style. The dataset is composed of sentences from the domain of medical science: 130,349 sentences were written in medical expert style and 114,674 sentences in layman style. A small subset of the dataset contains aligned sentence pairs in both styles.

\begin{table*}
\centering
\begin{tabular}{lccccll}
\hline
\textbf{Dataset Name} & \textbf{Year} & \textbf{Parallel (Y/N)} & \textbf{Type of Text Samples} & \textbf{Number of Samples} & \textbf{Labels for Style} & \textbf{References} \\
\hline
Enron corpus~\cite{cheng2020contextual, klimt2004introducing} & 2004 & Y & emails & 600K & \shortstack[l]{informal \\ formal} & \cite{cheng2020contextual} \\
\hline
Shakespeare~\cite{xu2012paraphrasing, xu2014data} & 2012 & Y & sentences & 21K & \shortstack[l]{shakespearean english \\ contemporary english} & \cite{zhao2018language} \\
\hline
\\
Gigaword~\cite{graff2003english, napoles2012annotated} & 2012 & N & news articles & 4M & seven publishers & \cite{zhang2018shaped} \\
\hline
SST~\cite{socher2013recursive} & 2013 & N & reviews & 9.6K & \shortstack[l]{positive \\ negative} & \cite{hu2017toward} \\
\hline
IMDB~\cite{diao2014jointly} & 2014 & N & reviews & 16K & \shortstack[l]{positive \\ negative} & \cite{dai2019style},~\cite{hu2017toward},~\cite{logeswaran2018content},~\cite{li2020dgst} \\
\hline
Amazon~\cite{he2016ups} & 2016 & N & reviews & 1M & \shortstack[l]{positive \\ negative} & \shortstack[l]{\cite{li2018delete},~\cite{sudhakar2019transforming},~\cite{fu2018style} \\ \cite{john2019disentangled},~\cite{liu2020revision},~\cite{wang2019controllable} \\ \cite{lee2020stable},~\cite{lample2018multiple},~\cite{kim2020positive}} \\
\hline
Gender~\cite{reddy2016obfuscating} & 2016 & N & reviews & $*$ & \shortstack[l]{female \\ male} & \cite{sudhakar2019transforming},~\cite{prabhumoye2018style},~\cite{prabhumoye2018style2} \\
\hline
Captions~\cite{gan2017stylenet} & 2017 & Y & image captions & 7K & \shortstack[l]{factual \\ romantic \\ humorous} & \cite{li2018delete},~\cite{sudhakar2019transforming},~\cite{wang2019controllable} \\
\hline
GYAFC~\cite{rao2018dear} & 2018 & Y & sentences & 110K & \shortstack[l]{informal \\ formal} & \cite{xu2019formality},~\cite{cheng2020contextual},~\cite{zhou2020exploring},~\cite{yi2020text} \\
\hline
Paper-News Titles~\cite{fu2018style} & 2018 & N & titles & 200K & \shortstack[l]{paper \\ news} & \cite{fu2018style},~\cite{duan2020pre} \\
\hline
Political slant~\cite{voigt2018rtgender} & 2018 & N & posts & 540K & \shortstack[l]{democratic \\ republican} & \cite{sudhakar2019transforming},~\cite{prabhumoye2018style},~\cite{prabhumoye2018style2},~\cite{tian2018structured} \\
\hline
Twitter~\cite{dos2018fighting} & 2018 & N & tweets & 2M & \shortstack[l]{offensive \\ non-offensive} & \cite{dos2018fighting} \\
\hline
Reddit~\cite{dos2018fighting} & 2018 & N & sentences & 7.5M & \shortstack[l]{offensive \\ non-offensive} & \cite{cheng2020contextual},~\cite{dos2018fighting} \\
\hline
Yelp & 2020 & N & reviews & 8.6M & \shortstack[l]{positive \\ negative} & \shortstack[l]{\cite{li2018delete},~\cite{sudhakar2019transforming},~\cite{dai2019style},~\cite{prabhumoye2018style} \\ \cite{john2019disentangled},~\cite{zhang2018learning},~\cite{prabhumoye2018style2},~\cite{tian2018structured} \\ \cite{liu2020revision},~\cite{wang2019controllable},~\cite{zhou2020exploring},~\cite{lee2020stable} \\ \cite{shen2017style},~\cite{zhao2018language},~\cite{logeswaran2018content},~\cite{chen2018adversarial} \\ \cite{lample2018multiple},~\cite{yi2020text},~\cite{huang2020cycle} \\ \cite{li2020dgst},~\cite{kim2020positive},~\cite{duan2020pre}} \\
\hline
Politeness~\cite{madaan2020politeness} & 2020 & N & emails & 270K & \shortstack[l]{neutral \\ polite} & \cite{madaan2020politeness} \\
\hline
Expertise~\cite{cao2020expertise} & 2020 & N & documents & 200K & \shortstack[l]{expertise \\ laymen} & \cite{cao2020expertise} \\
\hline
\multicolumn{7}{l}{} \\
\multicolumn{7}{l}{$*$ The dataset is composed of reviews written by 432M users, however the number of reviews is not specified by the authors.}
\end{tabular} 
\caption{\label{tab:datasets} A list of publicly available datasets for text style transfer.}
\end{table*}

\subsection{Evaluation of Automatic Style Transfer}

Evaluations of text style transfer face the longstanding challenges in the field of natural language generation (NLG)~\cite{gatt2018survey}. In regard to text style transfer, the objective of the evaluation is two-fold: 1) to measure how well the meaning of the original sentence was preserved in the output (generated sentence) and 2) to evaluate the quality of the style. Ideally, the goal is to create a model that successfully modifies the style of a text, while its meaning is preserved.

A different set of metrics have been used for evaluating both aspects. The quality of content preservation is evaluated using evaluation metrics that measure the extent to which the generated sentence matches human output, which is used in other NLG tasks, including summarization~\cite{chopra2016abstractive}, image captioning~\cite{vinyals2015show} and machine translation~\cite{bahdanau2015neural}. A new set of metrics, specifically tailored to measure the style strength, are proposed for measuring the quality of generating a text in the target style.

\subsubsection{Evaluation of the Quality of Semantic Content Preservation}

Despite the criticism of using metrics based on language modeling and similarity measures, a number of well-established metrics have been adopted for measuring the quality of text generation. 

Word overlap based metrics \textbf{METEOR}~\cite{denkowski2014meteor} and \textbf{BLEU}~\cite{papineni2002bleu}, were introduced for the evaluation of machine translation, by computing a score that indicates the similarity between the system output and one or more human-written reference texts. \textbf{METEOR}~\cite{denkowski2014meteor} evaluates the generated sentence by aligning it to one or more reference sentences. Alignments are based on exact, stem, synonym, and paraphrase match between words and phrases. METEOR is calculated as a harmonic mean of unigram precision and recall, with recall being weighted higher. \textbf{BLEU}~\cite{papineni2002bleu} measures how close a candidate sentence is to a reference sentence based on matches of n-grams of a sentence to a reference one. \textbf{NIST}~\cite{doddington2002automatic} is a version of BLEU metric that values the less frequent n-grams more. \textbf{BERTScore}~\cite{zhang2020bertscore} computes the cosine similarity between contextualized BERT~\cite{devlin2019bert} word embeddings of the sentence being evaluated and a set of reference sentences.

\textbf{ROUGE-L}~\cite{lin2004rouge} is a recall-oriented metric established for the evaluation of text summarization that applies the concept of the Longest Common Subsequence (LCS). The intuition behind the LCS concept is that the longer the LCS between two sentences, the more similar they are. The score is $1$ when the two sentences are equal, and $0$ when there is nothing in common between them.

\textbf{SARI}~\cite{xu2016optimizing} is a metric for text simplification that considers the number of additions and deletions. It measures the goodness of words that are added, deleted, and kept by the system. SARI first calculates precision and recall for each operation (addition, keep, and deletion). The final value is an average of these scores.

\textbf{PINC}~\cite{chen2011collecting} is a measure originally developed for evaluating paraphrasing. It evaluates how much a generated sentence resembles a reference sentence i.e. how many n-grams differ between the sentences. The final score is the percentage of n-grams that appear in the generated sentence but not in the reference. The novelty of paraphrases is greater as the value increases.

\subsubsection{Quality of a Style}

To evaluate the quality of generating a sentence in a target (output) style, various researchers calculate accuracy with a pre-trained classifier~\cite{li2018delete, sudhakar2019transforming, xu2019formality, dai2019style, cheng2020contextual, fu2018style, dos2018fighting, prabhumoye2018style, john2019disentangled, zhang2018learning, prabhumoye2018style2, liu2020revision, wang2019controllable, zhou2020exploring, lee2020stable, hu2017toward, shen2017style, zhao2018language, logeswaran2018content, chen2018adversarial}. A style quality is calculated as a percentage of generated sentences that were labeled with the target style by the classifier. Higher value indicates better style quality. Precision, recall, and F1-measure are also appropriate for the evaluation of the quality of a style.



\section{Deep Neural Networks for Text Generation}
\label{sec:deep_neural_networks_for_text_generation}

\subsection{Recurrent Neural Networks}

Recurrent Neural Networks (RNNs)~\cite{graves2013generating} are a class of deep learning networks designed for modeling sequential data. RNNs process the input sequence from beginning to end (forward direction). Bidirectional Recurrent Neural Networks (BiRNNs)~\cite{schuster1997bidirectional} are composed of two unidirectional RNNs operating in both directions (forward and backward).

Long Short-Term Memory Networks (LSTM)~\cite{hochreiter1997long} are a specific type of RNN, designed to learn long-range dependencies as well as to overcome the problem of vanishing and exploding gradients. Gated Recurrent Units (GRU)~\cite{chung2014empirical} have the same purpose as LSTM, but are known to be simpler and faster to train.

\subsection{Convolutional Neural Networks}

Convolutional Neural Networks (CNNs)~\cite{o2015introduction} are a class of deep neural networks developed for representing spatial features. Each neuron in CNNs is connected with a local region of neurons in the previous layer. The parameters are known as kernels that operate in two dimensions (2D convolution) over the spatial data thus producing two-dimensional feature maps of input. CNNs are suitable and are most commonly applied for image processing~\cite{krizhevsky2012imagenet, simonyan2014very}. Recently, CNNs have been explored for modeling sequential data~\cite{kalchbrenner2014convolutional, gehring2017convolutional, zhang2017sensitivity, zha2020gated, wang2021convolutional, adams2021private}, where the convolutional kernel operates in one dimension (1D convolution).

\subsection{Attention Mechanism}

Attention is a deep learning technique inspired by human cognitive attention that was introduced by~\citet{bahdanau2015neural} to improve machine translation. Attention is a technique, which computes a weighted sum of attention scores assigned to the elements of the input sequence that help the decoder to attend to certain elements of the input. There are various types of attention: dot-product attention, multiplicative attention, additive attention, self attention~\cite{luong2015effective, britz2017massive, vaswani2017attention}.

Self-attention is used for representation of a sequence while giving attention to relevant parts of the same sequence. Performing self-attention multiple times in parallel is defined as multi-head self-attention. Multi-head self-attention combines information from different representation subspaces. Transformer~\cite{vaswani2017attention}, a deep neural architecture proposed for language modeling by multi-head self-attention, allows significantly more parallelization than RNN. Various models have been built following the Transformer architecture: BERT~\cite{devlin2019bert}, DistilBERT~\cite{sanh2019distilbert}, RoBERTa~\cite{liu2019roberta}, GPT~\cite{radford2018improving}, GPT-2~\cite{radford2019language}, GPT-3~\cite{brown2020language}, etc.

Pointer network~\cite{vinyals2015pointer, merity2017pointer} is a mechanism for "pointing out" to relevant parts of the input. In a pointer network, attention is used as a pointer for selecting parts of the input sequence as members of the output sequence.

\subsection{Encoder-decoder}

Encoder-decoder (also referred to as sequence-to-sequence network) is a deep neural network architecture for text generation~\cite{sutskever2014sequence}. The encoder learns to generate a latent fixed-length vector representation of the input sentence. The decoder learns to generate an output sentence by decoding the fixed-length representation of the input sentence. The encoder and the decoder could be RNNs, CNNs, MLPs, attention-based networks, or a combination.

Autoencoder (AE)~\cite{rumelhart1985learning} and Variational Autoencoder (VAE)~\cite{kingma2014auto} are deep learning architectures intended for learning an internal representation of the input. As in the encoder-decoder architecture, the encoder is a neural network that produces fixed-length representation. The decoder learns to reconstruct the input sentence based on the encoded representation. VAE is a generative model that learns the distribution of the data with a stochastic variational and learning algorithm. AE and VAE could be viewed as a specific type of encoder-decoder architecture where the goal is to generate an encoded representation of the input data.

Encoder-decoder network has been applied in natural language generation for a number of tasks: machine translation~\cite{cho2014properties, wu2016google, bahdanau2015neural, cho2014learning}, text summarization~\cite{shi2021neural}, question answering~\cite{liu2020neural}, etc.

\subsection{Generative Adversarial Networks}

Generative Adversarial Network (GAN)~\cite{goodfellow2014generative} is a deep neural network architecture comprised of two networks – generator and discriminator. These two networks are trained simultaneously in a two-player minimax game. The generator network aims to learn the distribution of the training data and to generate samples from the learned distribution. The discriminator network determines whether a sample is from the data distribution or from the model distribution, by maximizing the probability of assigning the correct label to samples from the training data as well as from the generated data. The objective of the generator is to generate samples that are indistinguishable from the training samples, by maximizing the opposite objective of the discriminator. GANs have been applied in many natural language generation tasks~\cite{haidar2019textkd, ahamad2019generating, bergestext} including machine translation~\cite{wu2018adversarial}, text summarization~\cite{liu2018generative} and question answering~\cite{rao2019answer}.

\section{Methods for Style Transfer}
\label{sec:methods_for_style_transfer}

In this section, the discussion of deep learning (DL) models that were used in the research on text style transfer is presented. The vast majority of the models are built upon the encoder-decoder architecture~\cite{li2018delete, sudhakar2019transforming, xu2019formality, dai2019style, cheng2020contextual, fu2018style, dos2018fighting, prabhumoye2018style, john2019disentangled, zhang2018learning, prabhumoye2018style2, tian2018structured, liu2020revision, wang2019controllable, zhou2020exploring, lee2020stable, zhang2018shaped, hu2017toward, lample2018multiple, kim2020positive}, while in another line of research, adversarial learning with GANs have been put forward~\cite{shen2017style, zhao2018language, logeswaran2018content, chen2018adversarial, li2020dgst, yi2020text, duan2020pre, huang2020cycle}.

\begin{figure*}[ht]
\centering
\includegraphics[width=0.95\textwidth]{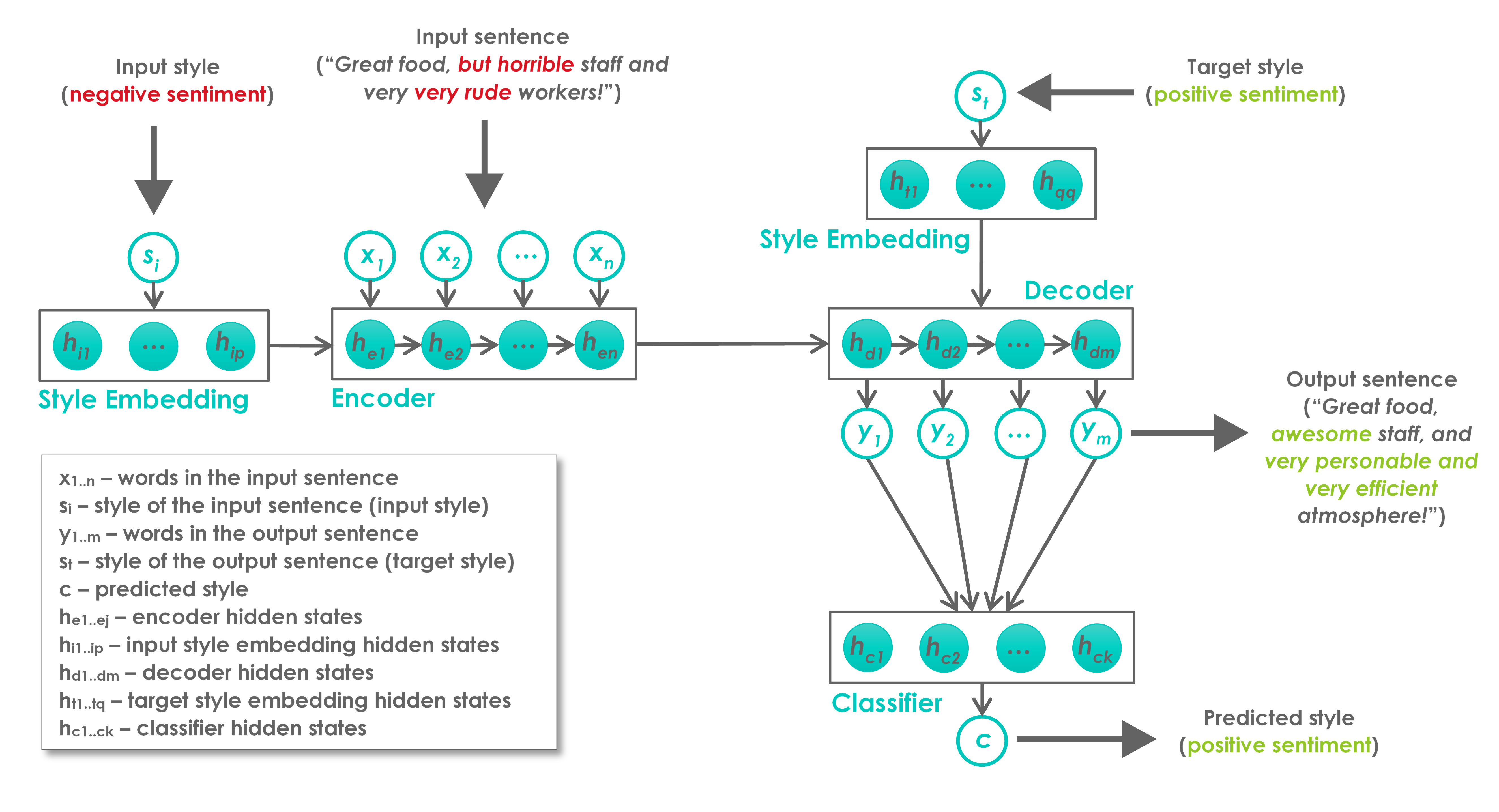}
\caption{General architecture of deep neural style transfer model.} \label{fig:general_architecture}
\end{figure*}

A general encoder-decoder-based architecture for style transfer is depicted in Figure~\ref{fig:general_architecture}. The encoder creates a latent representation of the input sentence i.e., encodes the input sentence. The decoder generates the output sentence conditioned on the latent representation, while the classifier determines the style label of the output sentence. The classifier is an optional component in style transfer models. If the style transfer model is based on GAN, the encoder is referred to as generator and the classifier component as a discriminator. Style embeddings have been introduced to assist in the encoding of input sentence and/or generation of output sentence. Style embedding could be added as input to the encoder~\cite{lee2020stable, hu2017toward}, decoder~\cite{li2018delete, sudhakar2019transforming, cheng2020contextual, fu2018style, zhang2018learning, logeswaran2018content, lample2018multiple, kim2020positive}, or both~\cite{dai2019style, dos2018fighting, shen2017style, chen2018adversarial}.

\subsection{Representation Learning}

An encoder is applied to create a latent vector representation of the content of a sentence. It transforms a sentence while preserving its semantic and syntactic properties. The initial building block, the encoder in style transfer models is usually a type of RNN. Two types of RNNs, LSTM~\cite{zhang2018learning, liu2020revision, hu2017toward, chen2018adversarial, li2020dgst, lample2018multiple, huang2020cycle} and GRU~\cite{li2018delete, fu2018style, dos2018fighting, john2019disentangled, tian2018structured, zhou2020exploring, shen2017style, logeswaran2018content}, have been employed to learn the representation of input sentences. 

In the literature on text style transfer, two types of encoders can be identified: shared and private encoders. When a shared encoder is used, the parameters are shared across the sentences of the entire dataset, so the encoder learns the style characteristics of the entire dataset. A private encoder is used to learn style-specific characteristics since the parameters are shared only across sentences of a specific style. Most of the style transfer models have opted for a shared encoder.~\citet{zhang2018shaped} have introduced a system that uses both, private and shared encoders. Each sentence is passed through two GRU encoders, one private encoder for a particular style and one encoder shared across all styles.~\citet{zhao2018language} proposed decomposing each sentence into two latent representations by using two GRU encoders, one for style representation, and the other for creating content representation. In a model called StyIns~\cite{yi2020text}, instead of learning style embeddings from a single sentence, the generative flow techniques~\cite{rezende2015variational} have been used to learn the stylistic properties from a set of sentences sharing the same style i.e., style instances. Coupled with an attention-based decoder, StyIns model yielded higher style accuracy while preserving the content of the original sentence when evaluated on three style transfer tasks.

Motivated by the findings that architectures for machine translation preserve the semantic meaning of a sentence, but not its stylistic properties~\cite{rabinovich2017personalized}, ~\citet{prabhumoye2018style, prabhumoye2018style2} have incorporated a machine translation-based model in the first stage of style transfer i.e., for representation learning of input sentences. While deep learning architectures for machine translation have reached state-of-the-art performances for many languages, their integration into deep learning pipelines for other tasks still face challenges.

Various models for text style transfer employ variants of Transformer architecture~\cite{xu2019formality, dai2019style, cheng2020contextual, wang2019controllable, lee2020stable, kim2020positive}, to benefit from the self-attention mechanisms when learning the representation of the input sentence.

An additional step in the process of style transfer, related to detection and removal of style markers from the input sentence, has been included in a number of studies~\cite{li2018delete, sudhakar2019transforming, zhang2018learning, lee2020stable}. Style markers are words that have the most discriminative power for determining the style of a sentence. Models proposed by~\citet{sudhakar2019transforming} do not use an encoder to create a latent representation of the sentence. Instead, the input sentence is reduced by removing the style markers and then it is passed directly to the decoder.

\subsubsection{Detecting Style Markers}

In most models for style transfer, the entire sentence is fed into the model~\cite{xu2019formality, dai2019style, cheng2020contextual, fu2018style, dos2018fighting, prabhumoye2018style, john2019disentangled, prabhumoye2018style2, tian2018structured, liu2020revision, wang2019controllable, zhou2020exploring, zhang2018shaped, hu2017toward, shen2017style, zhao2018language, logeswaran2018content, chen2018adversarial}. ~\citet{li2018delete},~\citet{sudhakar2019transforming},~\citet{zhang2018learning} and~\citet{lee2020stable} argue that style transfer can be accomplished by changing a few style markers. Based on this idea, the input sentence is preprocessed in a way that style markers are removed from it. The preprocessed sentence is then fed into the model as an input sentence. It should be noted that detecting and removing style markers has been proposed and evaluated only on the style transfer task of sentiment modification.

Several approaches for detecting style markers have been proposed.~\citet{li2018delete} have used n-gram salience measure for identifying style markers. It calculates the relative frequency of n-grams in sentences with a specific style. An n-gram is considered to be style marker if its salience is above a specific threshold.~\citet{zhang2018learning} used attention weights~\cite{kim2017structured} to detect style markers. A word is defined as a style marker, if its attention weight is greater than the average attention value.~\citet{sudhakar2019transforming} introduced an importance score for each token in the input sentence, based on attention scores of BERT~\cite{devlin2019bert} style classifier. Tokens with the highest importance score represent style markers.~\citet{lee2020stable} proposed to identify style markers by monitoring the change in probabilities of a style classifier. Important Score (IS) of a token is defined as a difference between the probability of the style conditioned on the entire sentence and the probability of the style conditioned on a sentence without the specific token. Token is a style marker if it has largest IS.

\subsection{Sentence Generation}

A crucial component of deep style transfer models is the part that generates the output sentence based on the representation of the content of the original sentence and the corresponding styles. We start our discussion with two models for sentiment modification, proposed by~\citet{li2018delete}, that are often used as baseline methods other research is compared to. The two models are not based on deep learning, but they use rather simplistic methods of retrieval or word swapping using two corpuses of sentences with positive and negative sentiment. The \textbf{RetrieveOnly} model simply outputs the retrieved sentence from the corpus that is most similar to the input sentence. A grammatically correct output sentence is expected, although the content of the original sentence might not be preserved. 

The \textbf{TemplateBased} model removes the words identified as style markers from the input sentence and replaces them with the style markers from the retrieved corpus sentence that is most similar to the original. This is a naïve method of word swapping based on the assumption that original style markers could be replaced with words with the opposite sentiment if they appear in a similar context (retrieved sentence). It is not surprising that very often the generated output sentence appears to be grammatically incorrect.

The first deep learning models developed for text style transfer were inspired by sequence-to-sequence systems for machine translation~\cite{bahdanau2015neural} and paraphrasing~\cite{prakash2016neural, cao2017joint, gupta2017deep}. An MT-based sequence-to-sequence model has been proposed for transforming text from modern English to Shakespearean English~\cite{jhamtani2017shakespearizing}. The output probability distribution is produced by a two-part decoder, consisting of an attention-based LSTM decoder and a pointer network. The pointer network was added to facilitate direct copying of the words from the input into the output sequence. Pre-trained word embeddings using external sources, such as dictionaries have been used to mitigate the problem of a limited amount of parallel data. An interesting approach of harnessing rules for formality style transfer using pre-trained GPT-2 Transformers was incorporated in several models proposed by~\citet{wang2019harnessing} to overcome the problem of small parallel datasets.

We group the models for style transfer into three groups according to the architectural blocks for generating the output sentence: \textit{simple reconstruction models}, \textit{models with style classifiers}, and \textit{adversarial models}. Simple reconstruction models are trained to simply reconstruct the sentences based on the reconstruction loss (called self-reconstruction). Models in the second group, incorporate additional style classifier(s) to assist in the generation of the output sentences, while models in the adversarial group are built upon GAN architecture.

\subsubsection{Simple Reconstruction Models}

Most of the style transfer models generate the output sentence as a simple “reconstruction” of the content of the input sentence by maximizing the probability distribution of the next word in the output sequence conditioned on the latent representation of the words in the input sentence. Style transfer research that belongs to the group of simple reconstruction models are presented in Table~\ref{tab:methods_reconstruction}. 

Two of the pioneering encoder-decoder models using simple reconstruction, \textbf{DeleteOnly} and \textbf{DeleteAndRetrieve}~\cite{li2018delete} have been proposed by the authors advocating style markers removal for the sentiment modification task. The decoder of the \textbf{DeleteOnly} model generates the output sentence based on the encodings of the input sentence and the target style (encoded by a separate style encoder). The style of the retrieved corpus sentence that is most similar to the input sentence is encoded in the \textbf{DeleteAndRetrieve} model to condition the output sentence generation.

\citet{sudhakar2019transforming} proposed two models, Blind Generative Style Transformer (\textbf{B-GST}) and Guided Generative Style Transformer (\textbf{G-GST}), that follow the same modeling approach as \textbf{DeleteOnly} and \textbf{DeleteAndRetrieve}~\cite{li2018delete}, respectively. However, both models incorporate Transformer-based decoders.

\citet{lample2018multiple} have suggested using a back-translation technique instead of adversarial training in their model \textbf{MultipleAttrTransfer}. By doing this, the generated output by the model is also used as a training input example fed into the encoder during “back-translation”. In terms of the objective function being optimized, a so-called cycle reconstruction is added to the original denoising auto-encoder. In addition, by using a temporal max-pooling layer on top of the encoder i.e. latent representation pooling, the decoder has better control over content preservation.

Shared-Private Encoder-Decoder (\textbf{SHAPED})~\cite{zhang2018shaped} is composed of multiple GRU encoders and multiple GRU decoders to learn both general and style-specific characteristics. One encoder and one decoder are shared across all styles in addition to the private encoders and decoders for each style. The outputs of both, private and shared encoder-decoders are concatenated and processed with a multi-layer feed-forward network to generate the output sentence. When the style of the input sentence is unknown, the sentence is fed into all private encoders. Their outputs are concatenated and fed into a style classifier that determines the style of the input sentence.

\citet{zhang2018learning} take on another approach to the task of sentiment modification in their Sentiment-Memory based autoencoder (\textbf{SMAE}). A sentiment classifier with self-attention mechanism is utilized to separate sentiment from non-sentiment words, creating “sentiment memories” i.e. weighted matrices of positive and negative sentiment word embeddings. During decoding, the context of the input sentence is used to extract closely-related sentiment entries from the sentiment memory matrix to condition the output sentence generation

\begin{table*}
\centering
\begin{tabular}{lclcl}
\hline
\textbf{Model} & \textbf{Year} & \textbf{DL Architecture} & \textbf{Style Transfer Task(s)} & \textbf{Dataset(s)} \\
\hline
RetrieveOnly~\cite{li2018delete} & 2018 & Baseline ML model & Sentiment Style Transfer & Yelp, Amazon, Captions \\
\hline
TemplateBased~\cite{li2018delete} & 2018 & Baseline ML model & Sentiment Style Transfer & Yelp, Amazon, Captions \\
\hline
DeleteOnly~\cite{li2018delete} & 2018 & \shortstack[l]{GRU encoder \\ GRU decoder} & Sentiment Style Transfer & Yelp, Amazon, Captions \\
\hline
DeleteAndRetrieve~\cite{li2018delete} & 2018 & \shortstack[l]{GRU encoder \\ GRU decoder} & Sentiment Style Transfer & Yelp, Amazon, Captions \\
\hline
SHAPED~\cite{zhang2018shaped} & 2018 & \shortstack[l]{multiple GRU encoders \\ multiple attentive GRU decoders} & Genre Transfer & Gigaword \\
\hline
SMAE~\cite{zhang2018learning} & 2018 & \shortstack[l]{LSTM encoder \\ LSTM decoder} & Sentiment Style Transfer & Yelp \\
\hline
B-GST~\cite{sudhakar2019transforming} & 2019 & BERT decoder & \shortstack[c]{Sentiment Style Transfer \\ Personal Style Transfer} & \shortstack[l]{Yelp, Amazon, Captions \\ Political slant, Gender} \\
\hline
G-GST~\cite{sudhakar2019transforming} & 2019 & BERT decoder & \shortstack[c]{Sentiment Style Transfer \\ Personal Style Transfer} & \shortstack[l]{Yelp, Amazon, Captions \\ Political slant, Gender} \\
\hline
MultipleAttrTransfer~\cite{lample2018multiple} & 2019 & \shortstack[l]{LSTM encoder \\ attentive LSTM decoder} & Sentiment Style Transfer & Yelp, Amazon \\
\hline
\end{tabular}
\caption{\label{tab:methods_reconstruction} Selection of style transfer research using simple reconstruction of the input sentences in a new style.}
\end{table*}

\subsubsection{Models with Style Classifier}

A large group of models incorporate a style classifier to facilitate the generation of the output sentence. The group of models shown in Table~\ref{tab:methods_classification} incorporate a style classifier in the process of generating the output sentence. 

\textbf{ControllableAttrTransfer}~\cite{wang2019controllable} first embeds the input sentence with a Transformer encoder, and then generates the output sentence with a Transformer decoder. The encoded representation is additionally fed into a two-layer linear classifier to provide a direction for editing the latent representation, so that it conforms to the target style.

Back-translation for Style Transfer (\textbf{BST})~\cite{prabhumoye2018style} is a back-translation based model that learns to rephrase the sentence by reducing the effects of the original style using English-to-French machine language translation model. A style classifier is used to identify the style of the latent representation of the back-translation model that later guides the generation of the output in style-specific generators. \textbf{BST} includes multiple BiLSTM style-specific decoders, one for each style. A style classifier is used to identify the style of the latent representation of the back-translation model that later guides the generation of the output sentence by the style-specific decoders. Multi-lingual Back-translated Style Transfer (\textbf{M-BST}) and Multi-lingual Back-translated Style Transfer + Feedback (\textbf{M-BST+F})~\cite{prabhumoye2018style2} are extensions of \textbf{BST}. \textbf{M-BST} creates latent sentence representation with multilingual MT model, while \textbf{BST} exploits monolingual MT model. In \textbf{M-BST+F}, a feedback-based loss function was introduced to guide the decoder.

Neural Text Style Transfer (\textbf{NTST})~\cite{dos2018fighting} and \textbf{StableStyleTransformer}~\cite{lee2020stable} utilize a CNN classifier to classify the style of the generated sentence. Unlike \textbf{BST}, these models are composed of a single decoder and therefore the inclusion of a style embedding as input to the decoder is needed to generate a sentence in the desired style. 

\textbf{StyleTransformer}~\cite{dai2019style} model uses a discriminator network as another Transformer encoder to distinguish the styles of sentences. The authors have experimented with two types of discriminator networks: a conditional discriminator to confirm or not the input style, and a multi-class discriminator that classifies a given sentence to one of the K style classes. The training algorithm goes through two phases: one for training the discriminator and the other is for training of the StyleTransformer network. For better preservation of the content of the original style, a cycle reconstruction loss is used as an objective function when training the model to generate the original input sentence if the generated output sentence is fed into the network. 

Context-aware Style Transfer (\textbf{CAST})~\cite{cheng2020contextual} trains two separate decoders for each sentence to ensure coherence with the adjacent context i.e., sentences in the same paragraph. By doing this, the model was able to preserve the style-independent content of the input sentence, while maintaining its consistency with the surrounding text.

To make use of parallel dataset when available, a bidirectional translation loss was introduced in \textbf{HybridST}~\cite{xu2019formality}, which is a combination of: 1) the loss of generating the output sentence given the input sentence and 2) the loss of generating an input sentence given an output sentence. In Fine-Grained Controlled Text Generation (\textbf{FineGrainedCTGen})~\cite{liu2020revision} additional Bag of Words (BOW) component was added to enhance generation of specific words and to preserve the content by minimizing the negative log probability of generating BOW features for the output sentence.

In \textbf{POS-LM}~\cite{tian2018structured}, two additional components were added: Part of Speech (POS) tagger and Language Model (LM). POS tagger assists in generating previously determined nouns in the output sentence, while LM controls the perplexity of the generated sentence.~\citet{zhou2020exploring} point out that training a model with reconstruction and classification loss might result in extremely short sentences that might match the target style, but would fail to preserve the original meaning. They proposed the \textbf{CP-LM} model that incorporates two losses: 1) content preservation loss to force the word embedding representation of the input and output sentences to be close, by minimizing the difference of their embedding representations and 2) fluency modeling loss to ensure that the output sentences are fluent by minimizing the negative log probability of generated words, similar to the bidirectional translation loss used in \textbf{HybridST}.

~\citet{kim2020positive} argue that performing sentence reconstruction and style control in a single task increases the complexity of the model. Their proposed model \textbf{AdaptiveStyleEmbedding} consists of two modules, one for each task: 1) a style module that learns the style embeddings using a style classifier, and 2) an autoencoder that generates the output sentence conditioned on the combined vector of latent representation of the input sentence and the learned style embedding.

\begin{table*}
\centering
\begin{tabular}{lclcl}
\hline
\textbf{Model} & \textbf{Year} & \textbf{Architecture} & \textbf{Style Transfer Task(s)} & \textbf{Dataset(s)} \\
\hline
BST~\cite{prabhumoye2018style} & 2018 & \shortstack[l]{MT encoder \\ multiple BiLSTM decoders \\ CNN classifier} & \shortstack[c]{Sentiment Style Transfer \\ Personal Style Transfer} & Yelp, Gender, Political slant \\
\hline
M-BST~\cite{prabhumoye2018style2} & 2018 & \shortstack[l]{multilingual MT encoder \\ multiple BiLSTM decoders \\ CNN classifier} & \shortstack[c]{Sentiment Style Transfer \\ Personal Style Transfer} & Yelp, Gender, Political slant \\
\hline
M-BST+F~\cite{prabhumoye2018style2} & 2018 & \shortstack[l]{multilingual MT encoder \\ multiple BiLSTM decoders \\ CNN classifier} & \shortstack[c]{Sentiment Style Transfer \\ Personal Style Transfer} & Yelp, Gender, Political slant \\
\hline
NTST~\cite{dos2018fighting} & 2018 & \shortstack[l]{GRU encoder \\ attentive GRU decoder \\ CNN classifier} & \shortstack[l]{Transferring Offensive \\ to Non-offensive Text} & Reddit, Twitter \\
\hline
POS-LM~\cite{tian2018structured} & 2018 & \shortstack[l]{GRU encoder \\ attentive GRU decoder \\ CNN classifier} & \shortstack[c]{Sentiment Style Transfer \\ Personal Style Transfer} & Yelp, Political slant \\
\hline
HybridST~\cite{xu2019formality} & 2019 & \shortstack[l]{Transformer encoder \\ Transformer decoder \\ CNN classifier} & Formality Transfer & GYAFC \\
\hline
StyleTransformer~\cite{dai2019style} & 2019 & \shortstack[l]{Transformer encoder \\ Transformer decoder \\ Transformer discriminator} & Sentiment Style Transfer & Yelp \\
\hline
ControllableAttrTransfer~\cite{wang2019controllable} & 2019 & \shortstack[l]{Transformer encoder \\ Transformer decoder \\ MLP classifier} & Sentiment Style Transfer & Yelp, Amazon, Captions \\
\hline
CP-LM~\cite{zhou2020exploring} & 2020 & \shortstack[l]{GRU encoder \\ attentive GRU decoder \\ CNN classifier} & \shortstack[c]{Sentiment Style Transfer \\ Formality Transfer} & Yelp, GYAFC \\
\hline
CAST~\cite{cheng2020contextual} & 2020 & \shortstack[l]{Transformer encoder \\ Transformer decoder \\ CNN classifier} & \shortstack[c]{Formality Transfer \\ Transferring Offensive \\ to Non-offensive Text} & GYAFC, Enron corpus, Reddit \\
\hline
StableStyleTransformer~\cite{lee2020stable} & 2020 & \shortstack[l]{Transformer encoder \\ Transformer decoder \\ CNN classifier} & Sentiment Style Transfer & Yelp, Amazon \\
\hline
FineGrainedCTGen~\cite{liu2020revision} & 2020 & \shortstack[l]{LSTM encoder \\ LSTM decoder \\ MLP classifier} & Sentiment Style Transfer & Yelp, Amazon \\
\hline
AdaptiveStyleEmbedding~\cite{kim2020positive} & 2020 & \shortstack[l]{Transformer encoder \\ Transformer decoder \\ MLP classifier} & Sentiment Style Transfer & Yelp, Amazon \\
\hline
\end{tabular}
\caption{\label{tab:methods_classification} Selection of style transfer research using a classifier to assist in the generation of sentences.}
\end{table*}

\subsubsection{Adversarial Models}

Several style transfer models incorporate style discriminators, which have a similar role as the discriminator in the GAN architecture. The research studies using the adversarial framework for generating the output sentence are listed in Table~\ref{tab:methods_adversarial}.

\textbf{AttrControl}~\cite{logeswaran2018content} is an encoder-decoder based model that employs a Projection Discriminator~\cite{miyato2018cgans} to generate realistic and style compatible sentences. The discriminator determines whether the generated sentence is real or fake, based on a style embedding and output sentence obtained by a GRU decoder. Controlled Text Generation (\textbf{CTGen})~\cite{hu2017toward} is built upon VAE architecture. Additional CNN discriminators were included to assist the generation process. The generator and the discriminators provide feedback to each other in a collaborative manner with the wake-sleep procedure~\cite{hinton1995wake}.

Aligned Autoencoder (\textbf{AAE})~\cite{shen2017style} incorporates a feed-forward discriminator to align both, posterior probability distributions learned with encoders for each style (input and target style). Cross-Aligned Autoencoder (\textbf{CAAE})~\cite{shen2017style} incorporates two CNN discriminators for the same purpose. Assuming the transfer is between two styles (style $s_i$ and style $s_t$), one discriminator learns to distinguish between a real sentence with style $s_i$ and generated sentence with style $s_t$, while the other discriminator learns to distinguish between a real sentence with style $s_t$ and generated sentence with style $s_i$.

\textbf{MultiDecoder} and \textbf{StyleEmbedding}~\cite{fu2018style}, incorporate two multi-layer classifiers to classify the style of the input sentence given the representation learned by the GRU encoder, by 1) maximizing the probability of correctly predicting style labels, and 2) maximizing the entropy of the predicted style labels. \textbf{StyleEmbedding} uses an additional embedded representation of the target style to the GRU decoder to generate an output sentence, while \textbf{MultiDecoder} is composed of multiple GRU decoders (one for each style) to generate a sentence in the target style. 

The research study presented by~\citet{john2019disentangled} tackled on somewhat divisive topic of the feasibility of disentangling the content from the style in the latent space. Deterministic AutoEncoder (\textbf{DAE}) and Variational AutoEncoder (\textbf{VAE}) have been used in their models for the task of sentiment style transfer. A number of content-oriented and style-oriented reconstruction and adversarial losses have been proposed to afford the separation of the latent spaces. The content and style information have been approximated by the bag-of-words (BOW) features. Two classifiers have been used: one for detecting the style and the other over the BOW content vocabulary.

Cycle-consistent Adversarial Autoencoder (\textbf{CAE})~\cite{huang2020cycle} is a three-component network consisting of LSTM autoencoder for representing sentences in different styles, adversarial style transfer network, and a novel cycle-consistent constraint. The cycle-consistent reconstruction imposes constraint on the latent representation collectively learned by the LSTM autoencoder and adversarial style network. The results of the conducted ablation study show that the cycle-constraint was instrumental in content preservation during sentiment style transfer. In a similar way, Dual-Generator Network for Text Style Transfer (\textbf{DGST})~\cite{li2020dgst} learns to generate sentences in a target style in a cyclic process. However, this model does not rely upon discriminators. Instead, it applies neighborhood sampling to introduce noise to each sentence. \textbf{FM-GAN}~\cite{chen2018adversarial} is trained with Feature Mover's Distance instead of traditional loss for adversarial learning.

~\citet{zhao2018language} point out that the generated sentence may not necessarily capture the target style by training with an objective function that includes only reconstruction and adversarial loss. Their model \textbf{StyleDiscrepancy} incorporates an additional discriminator to determine whether a given sentence has the target style with a loss function called style discrepancy. They also apply cycle consistency as in \textbf{StyleTransformer}~\cite{dai2019style} and \textbf{CAST}~\cite{cheng2020contextual} models. StyIns~\cite{yi2020text} model incorporates adversarial style loss to ensure better style supervision during generation. Similar to \textbf{StyleTransformer}~\cite{dai2019style}, a multi-class discriminator is applied to determine the style of the generated sentence.

A new framework named Pre-train and Plug-in Variational Autoencoder (\textbf{PPVAE}) was proposed by~\citet{duan2020pre} with a realistic system in mind that can mitigate the problem of starting from scratch whenever we need to learn a new style. The framework PPVAE is composed of two variational autoencoders: the PretrainVAE, which learns to represent and reconstruct a sentence in its original style and the PluginVAE, which learns the conditional latent space for each style. The role of PluginVAE as a lightweight easily-trained network is to transform the conditional “style-specific” latent space into the global latent space learned by PretrainVAE and vice versa.

\begin{table*}
\centering
\begin{tabular}{lclcl}
\hline
\textbf{Model} & \textbf{Year} & \textbf{Architecture} & \textbf{Style Transfer Tasks(s)} & \textbf{Dataset(s)} \\
\hline
AAE~\cite{shen2017style} & 2017 & \shortstack[l]{GRU encoder \\ GRU generator \\ MLP discriminator} & Sentiment Style Transfer & Yelp \\
\hline
CAAE~\cite{shen2017style} & 2017 & \shortstack[l]{GRU encoder \\ GRU generator \\ CNN discriminator} & Sentiment Style Transfer & Yelp \\
\hline
CTGen~\cite{hu2017toward} & 2017 & \shortstack[l]{LSTM encoder \\ LSTM generator \\ CNN discriminator} & Sentiment Style Transfer & SST, IMDB \\
\hline
AttrControl~\cite{logeswaran2018content} & 2018 & \shortstack[l]{GRU encoder \\ GRU decoder \\ Projection discriminator} & Sentiment Style Transfer & Yelp, IMDB \\
\hline
StyleDiscrepancy~\cite{zhao2018language} & 2018 & \shortstack[l]{two GRU encoders \\ GRU generator \\ CNN discriminator} & \shortstack[c]{Sentiment Style Transfer \\ Shakespearean Style Transfer} & Yelp, Shakespeare \\
\hline
FM-GAN~\cite{chen2018adversarial} & 2018 & \shortstack[l]{LSTM encoder \\ LSTM generator \\ MLP classifier} & Sentiment Style Transfer & Yelp \\
\hline
StyleEmbedding~\cite{fu2018style} & 2018 & \shortstack[l]{GRU encoder \\ GRU decoder \\ MLP classifier} & \shortstack[c]{Genre Transfer \\ Sentiment Style Transfer} & Paper-News Titles, Amazon \\
\hline
MultiDecoder~\cite{fu2018style} & 2018 & \shortstack[l]{GRU encoder \\ multiple GRU decoders \\ MLP classifier} & \shortstack[c]{Genre Transfer \\ Sentiment Style Transfer} & Paper-News Titles, Amazon \\
\hline
DAE~\cite{john2019disentangled} & 2019 & \shortstack[l]{GRU encoder \\ GRU decoder \\ CNN classifier} & Sentiment Style Transfer & Yelp, Amazon \\
\hline
VAE~\cite{john2019disentangled} & 2019 & \shortstack[l]{GRU encoder \\ GRU decoder \\ CNN classifier} & Sentiment Style Transfer & Yelp, Amazon \\
\hline
StyIns~\cite{yi2020text} & 2020 & \shortstack[l]{BiLSTM encoder \\ attentive RNN decoder \\ CNN discriminator} & \shortstack[c]{Sentiment Style Transfer \\ Formality Style Transfer} & Yelp, GYAFC \\
\hline
CAE~\cite{huang2020cycle} & 2020 & \shortstack[l]{LSTM encoders \\ LSTM generators \\ MLP discriminators} & Sentiment Style Transfer & Yelp \\
\hline
DGST~\cite{li2020dgst} & 2020 & \shortstack[l]{BiLSTM encoders \\ BiLSTM generators} & Sentiment Style Transfer & Yelp, IMDB \\
\hline
PPVAE~\cite{duan2020pre} & 2020 & \shortstack[l]{BiGRU and MLP encoders \\ Transformer and MLP decoders \\ MLP discriminator} & \shortstack[l]{Genre Transfer \\ Sentiment Style Transfer} & Yelp, News Titles \\
\hline
\end{tabular}
\caption{\label{tab:methods_adversarial} Selection of style transfer research using adversarial learning for generating sentences in a new style.}
\end{table*}

\section{Challenges and Directions for Future Research}
\label{sec:challenges}

Before concluding this paper, it is important to emphasize the challenges the task of text style transfer faces. Research performed thus far offers a promising perspective but also points to unexplored avenues that require further attention.

\subsection{Datasets}

Advancing the state-of-the-art systems for style transfer is dependent on the quality and quantity of the available datasets. They provide the necessary support for advocating the use of deep learning techniques. Due to the diversity of style categories, the ambiguity of the stylistic properties being modeled and the costs of labeling, creation of benchmarking datasets is still non-trivial. Publicly available datasets for style transfer vary depending on the type of style they contain (e.g., sentiment, formality, politeness) they contain, the format of text samples, and the procedure used for content validation and labeling (e.g., crowdsourcing, experts). While parallel corpus datasets would be desirable for style transfer tasks, it is often unrealistic to obtain and label large datasets, so the rise in unsupervised deep style transfer methods is not surprising.

\subsection{Deep Learning}

Heralded for their capabilities for automated feature learning and complex pattern recognition from vast quantities of big data, deep learning techniques have been at the frontier of innovations for decades now. The aim of this survey was to highlight the importance and to demonstrate the suitability of deep learning for the task at hand.

The focus of the research has shifted from feature extraction to model-free machine learning. The knowledge is unearthed directly from abundant data without the need for domain expertise, hand-crafted feature extraction, or data labeling. Deep learning is currently being preferred choice for text style transfer and have proved to be more scalable, robust, and superior in performance on various style transfer tasks. Existing deep neural architectures have been adapted for both stages in the process: the representation learning of the input sentence whose style needs to be changed and the generation of the output sentence in a new style. 

The technological solutions for output sentence generation were the criteria for clustering the style transfer research discussed in the survey into three groups. There are strengths and limitations associated with deep learning. Their dependency on large quantities of data and the complexity of the neural models are related to the problems of overfitting from training data and inability to generalize well. 

\subsection{Deep Style Transfer}

Points of particular interest for future directions in style transfer are expected in the following areas:

\subsubsection{Style-content Disentanglement}

The discussions that extend across several studies are the challenges in disentanglement of content and style in text. Style is indeed inseparably woven into spoken and written language. Some of the past research studies have advocated that the success of the style transfer task depends on the clear separation between the semantic content from the stylistic properties. Approaches being proposed include: partitioning of the latent space into content and style subspaces~\cite{fu2018style, john2019disentangled, hu2017toward, shen2017style, zhao2018language, yi2020text}, removal of style markers~\cite{li2018delete, sudhakar2019transforming, zhang2018learning, lee2020stable}, and use of back-translation for reducing the effects of the style of the input sentence~\cite{prabhumoye2018style, prabhumoye2018style2}.

Across recent studies, the majority of researchers clearly reject the necessity for disentangling content from style; a target that is difficult to reach even by adversarial training. A number of methods have been suggested that are much more effective on style transfer tasks without the need for disentanglement. The use of back-translation technique~\cite{dai2019style, cheng2020contextual, dos2018fighting, lample2018multiple}, latent representation editing~\cite{wang2019controllable}, independent modules for style control and sentence reconstruction~\cite{kim2020positive}, and a cycle-consistent reconstruction~\cite{huang2020cycle} are some of the approaches put forward. Given the promising success of adversarial-based method for style transfer task, there is evidence that further advancing the proposed adversarial architectures is worth exploring. Surely, these divisive views need to be further explored to consolidate the views across different style transfer tasks.

\subsubsection{Content Preservation vs. Style Strength Trade-off}

A key challenge for the methods for style transfer in text is identifying strategies that can effectively balance the trade-off between preserving the original content and changing the style of a given sentence~\cite{li2018delete, cheng2020contextual, liu2020revision, lee2020stable, li2020dgst, lample2018multiple, kim2020positive, yi2020text}. The balance is problematic because the precise nature of the interdependence between style-free and style-dependent content is not clearly defined. The unavoidable trade-off between content preservation and style strength in the proposed models is shared among researchers in the field. Furthermore, previous studies suggest that some technological remedies are at odds with one another~\cite{dai2019style, li2020dgst, huang2020cycle}. 

\subsubsection{Interpretability}

Deep neural networks are sometimes criticized for being black-box models, their structure and output not intelligible enough to associate causes with effects. In the field of natural language processing, we would want to interpret the model in a way that we can identify the useful patterns and features that contribute more to a better understanding and generation of text. Making attempts to understand how and how well different deep network components have been done in machine vision by dissecting GANs~\cite{bau2018gan}.

In the realm of natural language understanding, a number of perspectives has been fitted under the umbrella of style, from genre and formality to personal style and sentiment. This is an area where interdisciplinary endeavor of theoretical and empirical research should complement each other. The topic of stylistic language variations has a rich history in sociolinguistic theory~\cite{labov1972sociolinguistic, labov1981field, labov1972some}, although reciprocal contributions from both sides are expected to shed light on the multifaceted nature of style, provide guidance to the modeling efforts and improve the interpretation of the empirical results.

\subsubsection{Transfer Learning}

Training models is a computationally intensive and time-consuming process. In the field of computer vision and natural language understanding, transfer learning has been used to both, to speed up the process and improve model performance. General semantic patterns are learned during pre-training and can be "transferred" to new tasks. By fine-tuning, such additional semantic information can be "transferred" into the learned representation. Currently, the number of research studies using transfer learning for style transfer is still limited. However, as popularity and attention to the topic increases, the idea of pre-training, multi-task training, and then fine-tuning can be more promising.

\subsubsection{Ethical Considerations}

The far-reaching consequences of any research should engage the community in ethical discussions on potential malicious abuse as well as the impact a technology has on transforming many aspects of human life. The need for reflection is further amplified by the increased reliance of current machine learning technologies on abundant data that is scraped from social networks.

\subsubsection{Deep Reinforcement Learning}

The new directions towards human-like understanding, such as few-shot learning~\cite{wang2020generalizing}, inductive learning~\cite{murty2020expbert} or lifelong learning~\cite{chen2018lifelong} are promising areas of machine learning. A vision of designing machines that learn like humans from experience based on a limited number of training examples is yet to be reached~\cite{lake2017building}. Notably, deep reinforcement learning has recently been revisited by researchers in many fields including our own~\cite{li2016deep, li2018paraphrase, romain2018deep}.

The advances reviewed in this survey should be of interest to researchers not only limited to style transfer, but also to other related fields, such as language generation, summarization, question answering and dialogue. 

\section{Conclusion}
\label{sec:conclusion}

Recent advancements in text style transfer using deep learning have been the primary motivation to carry out the survey presented in this paper. A systematic review of state-of-the-art research highlights the trends that appeared to extend across research studies as well as differences and variations in style transfer methodologies using deep learning. In particular, this review examines how encoder-decoder-based architectures are still dominating the field, with a more recent move toward adversarial learning using Generative Adversarial Networks. While it appears that a choice of one deep neural network over another is style-independent, balancing the trade-offs between the complexity of a model and the expected performance gains added by auxiliary components (e.g., classifier, discriminator) are consistent challenges faced by researchers. The review is structured around the key stages in style transfer process and the methodological differences adopted by researchers for each stage.

It is our hope that the review would serve as a guideline for future studies that are built on the best practices of past research as well as the new direction that can enrich the field. Notably, successful studies of generalizing results across style transfer tasks are rarely reported. Transfer learning and multitask learning studies are the opportunities that could make further progress possible. Interpretability is a recurring challenge that is shared across respective fields – a better understanding of what stylistic indicators are captured and learned by neural models might elucidate the nature of stylistic variations in language.

\bibliography{main}
\bibliographystyle{IEEEtranN}


\end{document}